\documentclass{article}
\usepackage{spconf,amsmath,graphicx}
\usepackage{algorithm}
\usepackage[noend]{algpseudocode}
\def\algbackskip{\hskip-\ALG@thistlm}

\newcommand{\mycomment}[1]{}

\title{Attentive VQ--VAE}
\name{{Angello Hoyos and Mariano Rivera} 
\thanks{This work has been submitted to the IEEE for possible publication. Copyright may be transferred without notice, after which this version may no longer be accessible.}}
\address{Centro de Investigacion en Matematicas A.C. \\
         Guanajuato, Gto. 36120, Mexico}
\begin{document}
%
\maketitle
%
\begin{abstract}
We present a novel approach to enhance the capabilities of VQ-VAE models through the integration of a Residual Encoder and a Residual Pixel Attention layer, named Attentive Residual Encoder (AREN). The objective of our research is to improve the performance of VQ-VAE while maintaining practical parameter levels. The AREN encoder is designed to operate effectively at multiple levels, accommodating diverse architectural complexities. The key innovation is the integration of an inter-pixel auto-attention mechanism into the AREN encoder. This approach allows us to efficiently capture and utilize contextual information across latent vectors. Additionally, our models uses additional encoding levels to further enhance the model's representational power. Our attention layer employs a minimal parameter approach, ensuring that latent vectors are modified only when pertinent information from other pixels is available. Experimental results demonstrate that our proposed modifications lead to significant improvements in data representation and generation, making VQ-VAEs even more suitable for a wide range of applications as the presented.
\end{abstract}

\begin{keywords}
VQ--VAE, Attention, Face Generation, GANs.
\end{keywords}

\section{Introduction}
\label{sec:intro}

The field of generative models has seen significant advances in recent years, enabling the creation of high-quality and diverse synthetic data \cite{kingma2013auto,rezende2014stochastic,van2016pixel,van2017neural,razavi19vqvae2,ho2019flow++,chen2020generative,zhao2021vq}. Among these, the Variational Autoencoder (VAE) has emerged as a robust framework for learning latent representations of data distributions \cite{kingma2013auto}. However, traditional VAEs need help generating rich textures and tend to produce over-smoothed images. Thus, refinement models are used to improve the image's realness. More recently, the Variational Autoencoder with Vectored Quantization (VQ-VAE) \cite{razavi19vqvae2} was introduced to address VAE's limitations, offering a novel approach that combines the strengths of autoencoding with discrete vector quantization \cite{agustsson2017soft}. The VQ--VAE architecture provides a unique solution to the problem of capturing fine-grained details in data while maintaining the interpretability of latent representations. However, traditional VQ--VAE faces challenges in modeling complex dependencies and preserving long-range consistency in generated samples. With an extra computational cost, a solution consists of implementing a hierarchical codification \cite{razavi19vqvae2}. Hence, the latent vectors, $[{\bf z}]_{i=1,2,\ldots}$ in higher codification levels, are computed with an extended support region in the original image. 

Among applications of VQ-VAEs are image denoising, data compression \cite{agustsson2017soft}, data generation \cite{razavi2019generating,zhao2021vq}, abnormality detection \cite{marimont2021anomaly}, image/video super-resolution \cite{adiban2023s}, and denoising \cite{Peng_2021_CVPR} to mention a few. Herein, we introduce a the Attentive VQ--VAE (AREN) that incorporates a Attention mechanisms that extends the codification capabilities \cite{vaswani2017attention}. We succesfully compare our proposal with a Hierarchical variant \cite{zhao2020efficient}. These advancements enable the encoder to more effectively preserve intricate features in the generated samples. For instance, in image generation, Attentive VQ--VAE demonstrates improved capabilities in capturing subtle facial features, like the symmetry of facial attributes, color distribution of eyes, and nuanced contours of facial components. This paper presents the architecture and training strategy of Attentive VQ--VAE, demonstrating its effectiveness through extensive experimental results. Incorporating a Generative Adversarial Network (GAN) training strategy further enriches the model's capabilities by reducing the required training iterations. Our numerical experiments highlight the distinct advantages of Attentive VQ--VAE over its predecessors.

\begin{figure}[!t]
\centering
\includegraphics[width=\linewidth]{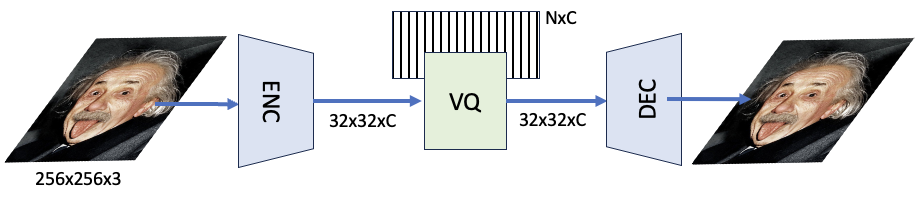}
\caption{General scheme of a VQ--VAE} 
\label{fig:vqvae}
\end{figure}

The remainder of this paper is organized as follows: Section 2 provides a comprehensive overview of VQ-VAE models. Section 3 details the architecture of the Attentive Hierarchical VQ--VAE, including its novel components and improvements over the original VQ--VAE framework. That Section also delves into the training strategy based on GANs techniques \cite{goodfellow2020generative}; indeed, a PatchGAN \cite{li2016precomputed}. Section 4 presents the experimental setup and showcases the results achieved by investigated versions of Attentive Hierarchical VQ--VAE, quantitatively and qualitatively comparing their performance. Section 4 also presents three application examples (blind restoration, denoising and deblurring) that demonstrate our model capabilities. Finally, Section 5 concludes the paper by summarizing the contributions and discussing potential future research directions in the Attentive VQ--VAE and generative modeling.

\section{Related Work}
\label{sec:related}

In Fig. \ref{fig:vqvae}, we depict the general scheme of the VQ-VAE. In this,  the encoder (ENC) transforms the input data from its original space into an array of latent vectors ${\bf z}$ (in dimension $N$) with minimal information loss. Subsequently, a Vector Quantizer (VQ) replaces each vector ${\bf z}_i \in {\bf z}$ with the vector 
\begin{equation}
{\bf e}_k = \arg\min_{{\bf e}_k \in D} \|{\bf z}_i-{\bf e}_k\|,    
\end{equation}
where $D$ is a dictionary of vectors learned from the data. Hence, each latent vector can only be one of those defined by the dictionary. In this architecture, such a dictionary is learned from the data at training time. Thus, if the dictionary $D$ exhibits sufficient diversity and the encoder effectively maps ${\bf z}$ to ${\bf e}$ with minimal error, then the decoder (DEC) is capable of generating a reliable reconstruction of the original data. An important distinction between the VQ-VAE and the traditional VAE lies in their treatment of latent variables: the former quantizes them \cite{maddison2017concrete, razavi19vqvae2}, while the latter constraint the latent variables by imposing them a prior distribution, often adopting a multivariate normal distribution with zero mean and identity covariance: $p({\bf z}) \sim \mathcal{N}(0, I)$.

VQ--VAE models have been noted for their efficiency in encoding data into low-dimensional latent spaces. Once one trains a VQ--VAE model, the prior distribution of the latent variables, $p({\bf z})$, can be learned using autoregressive models such as PixelCNN \cite{van2016pixel, van2016conditional}. Then, one can generate new data by sampling $p({\bf z})$ and decoding such samples. Our work focuses solely on improving the efficiency and capacity of VQ-VAE models. It is essential to mention that the prior distribution $p({\bf z})$ estimation is beyond our study's scope. We focus on the  VQ--VAE encoder's inherent loss of information challenge. In the context of facial images, VAE-based models have exhibited deficiencies in global consistency \cite{razavi2019generating,zhao2021vq}. These issues encompass generated faces with asymmetrical features (e.g., eyes of different colors) and overall incoherence (e.g., disproportional features). One can attribute these challenges to the latent vectors derived from convolutional networks, which capture characteristics from localized regions. However, enlarging the support region implies expanding convolutional kernels, leading to an escalation in parameters and training time. 

\section{Method}
\label{sec:proposal}

Our focus in this paper centers on enhancing the encoder's capabilities by redesigning encoders, reducing encoding resolution levels, and integrating an attention mechanism. These modifications aim to augment the performance of the VQ-VAE while maintaining parameters at practical levels.

\subsection{Attentive Residual Encoder}
\label{ssec:AREN}

\begin{figure}[!ht]
\centering
\includegraphics[width=\linewidth]{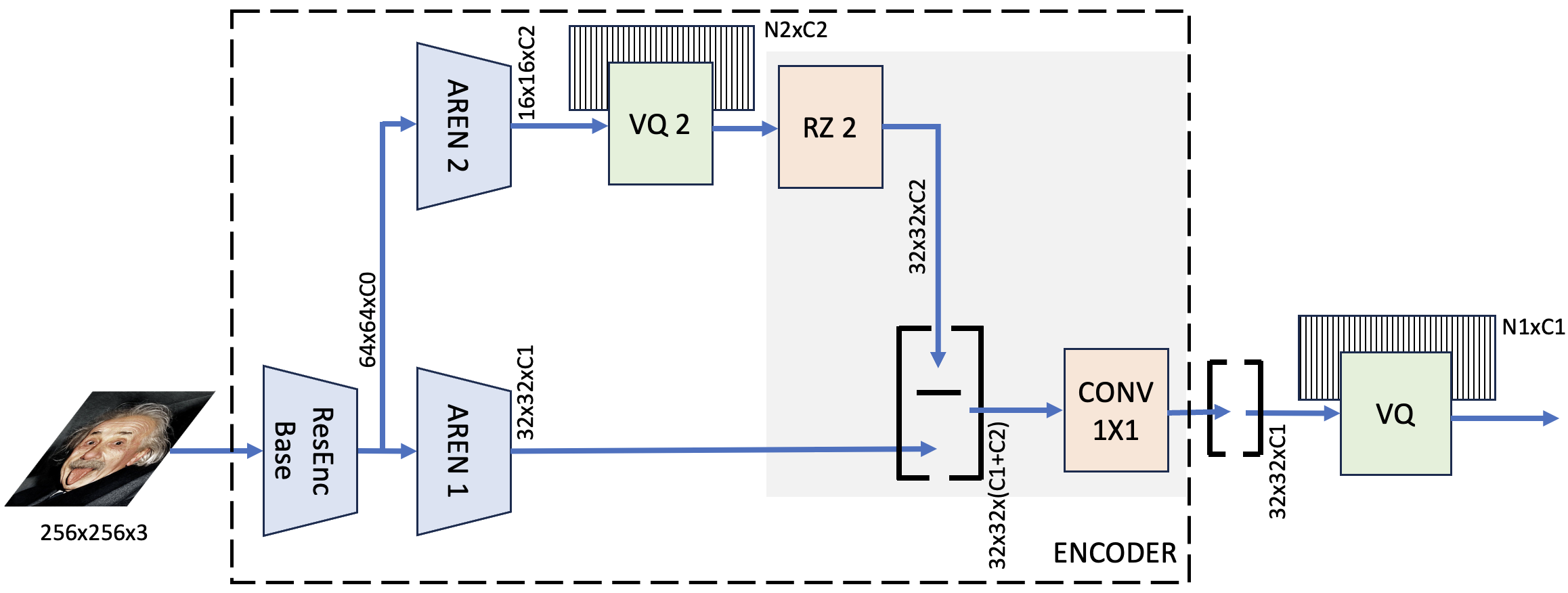}
\caption{Attentive Residual Encoder (AREN).} 
\label{fig:aren}
\end{figure}

Fig. \ref{fig:aren} depicts the schematic of our proposed Attentive Residual Encoder (AREN). Our encoder design draws inspiration from the multilevel encoder proposal of VQ-VAEv2. Although Fig. \ref{fig:aren} illustrates the two-level case, our implementation accommodates multiple levels (we have tested up to three). The dashed rectangle represents the encoder within the broader scheme, as shown in Fig. 1. Notably, the base encoder's output is shared among all encoder levels. We design the AREN-type encoders to operate complementary by effectively splitting the information corresponding to each level. The latent vector of the upper level undergoes quantization (VQ2) and scaling (RZ2) to align the Height and Width dimensions with those of the lower level. Then, we concatenate, by channels,  the quantized response of the upper level with the AREN response of the lower level. We combine the concatenated channels with a 1x1 convolution. This processed tensor serves as the output of the proposed encoder and is passed to the vector quantizer.

\begin{figure}[!ht]
\centering
\includegraphics[width=\linewidth]{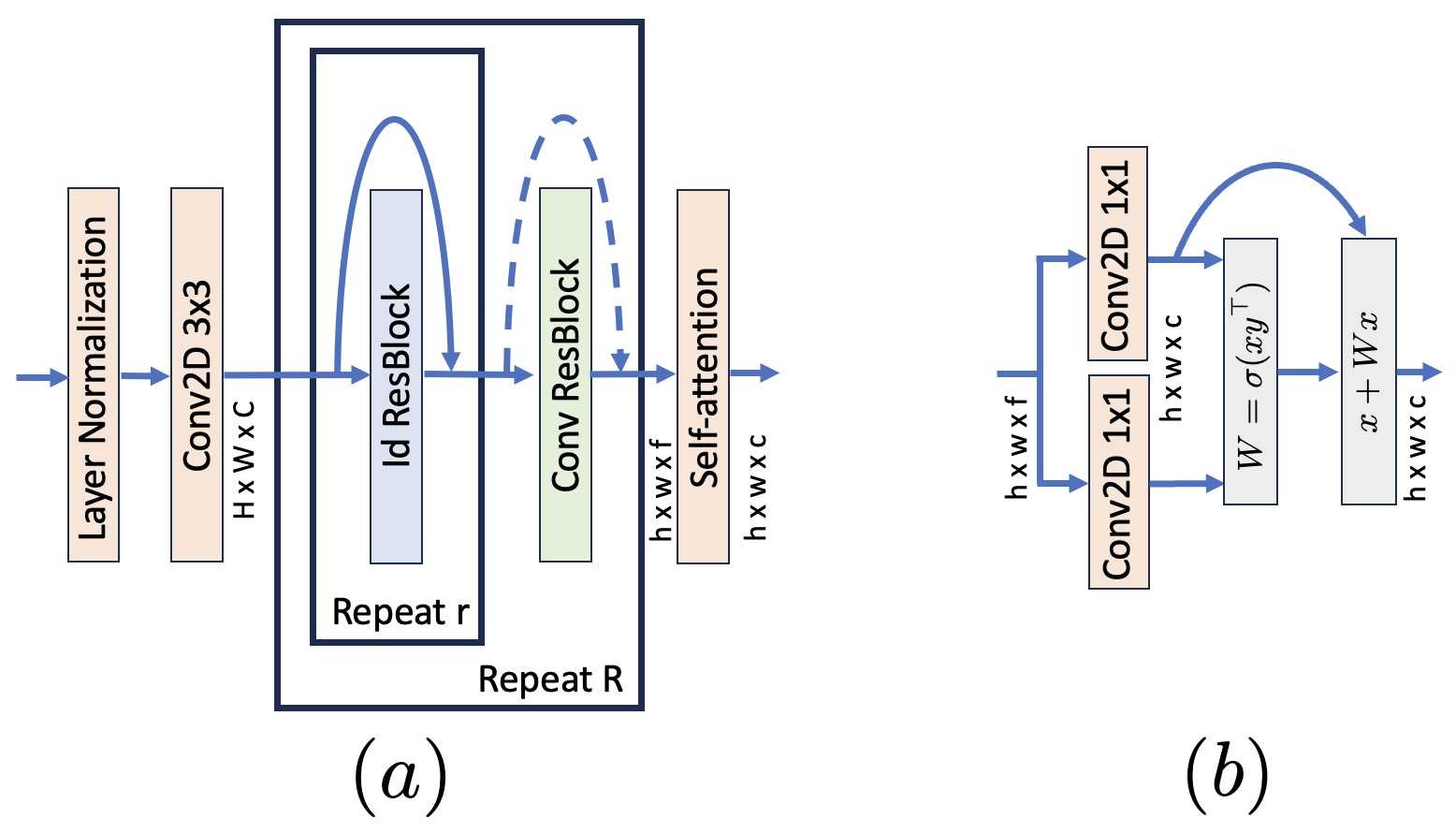}
\caption{Modules of AREN: (a) Residual Encoder, and (b) Self--attention module.} 
\label{fig:attention}
\end{figure}

In Panel (a) of Fig. \ref{fig:attention}, we illustrate the components of the AREN encoder. It builds upon the convolutional residual network architecture ResNetv2 \cite{he2016identity}, adding a pixel attention layer and a 1x1 convolution to adjust channel numbers. Particularly important is the inter-pixel auto-attention layer inspired by Pixel--Attention; see Panel (b) in Fig. \ref{fig:attention}. 

\noindent {\bf Additional Encoding Levels.} To introduce an extra encoding level, a hypothetical level 0 in Fig. \ref{fig:aren}, we take the output of the current lower level and adjust its dimensions to match the AREN output of the new lower level. Thus, concatenate by channels those tensors. The remaining aspects of the new lower level closely resemble those of the level immediately above it.

\subsection{Residual Pixel Attention}
\label{ssec:attention}

The attention layer objective is to incorporate information from similar pixels into each latent vector through a residual-based strategy. Our attention layer employs minimal parameters, and the residual approach ensures modifications to the latent vector only when pertinent information from other pixels is available.

Given the tensor $x$, with dimensions $(h,w,f)$,  the attention matrix $W$ is computed as
\begin{equation}
    W_{i j} = \sigma(g_1(x_i) \, g_2(x_j)^\top),
\end{equation}
where $g_1$ and $g_2$ are convolutional layers with kernel size equal to $1 \times 1$, $\sigma$ denotes the sigmoid activation function, and the number of filters equals the latent space dimension, $c$. Then, the residual attention layer implements:
\begin{equation}
    x \leftarrow x + W x    
\end{equation}
The pseudocode in Algorithm \ref{alg:PixelAttention} presents the details of the Attention layer. Such a pseudocode computes self-attention if we pass $y=x$ as parameters; in another case, a cross-attention.

\begin{algorithm}
\caption{PixelAttentionV2}\label{alg:PixelAttention}
\hspace*{\algorithmicindent} \textbf{Input}  Tensor $x$ with dimensions $(b,h,w,f)$ \\
\hspace*{\algorithmicindent} \textbf{Input}  Tensor $y : (b,h,w,f)$ \\
\hspace*{\algorithmicindent} \textbf{Output} Updated $x : (b,h,w,c)$ 
\begin{algorithmic}[1]
\Procedure{Attention}{$x,y$}
\State $y  \gets \textit{Conv2D}(c,1 \times 1)(y)$ \Comment{Num. filters $c$}
\State $x  \gets \textit{Conv2D}(c,1 \times 1)(x)$
\State $y \gets \textit{Reshape}(b,h \times w,c)(y)$
\State $x \gets \textit{Reshape}(b,h \times w,c)(x)$
\State $W_{b,i,j} \gets x_{b,i,c} \, y_{b,j,c}$ \Comment{Einstein notation (EN)}
\State $W \gets \sigma(W)$
\State $x_{b,i,c} \gets x_{b,i,c} + W_{b,i,j} \, y_{b,j,c}$  \Comment{EN}

\State $x \gets \textit{Reshape}(b,h,w,c)(x)$ \\
\Return $x$ 
\EndProcedure
\end{algorithmic}
\end{algorithm}

\begin{figure*}[!ht]
\centering
\includegraphics[width=\linewidth]{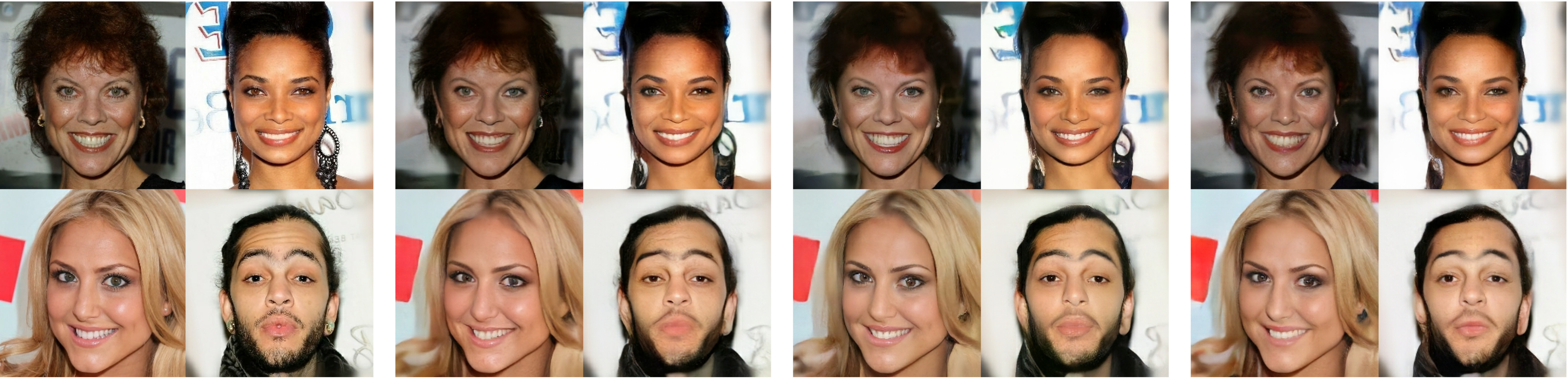}
\raggedright 
\text{\;\;\;\;\;\;\;\;\;\;\;\;\;\;\;\;\;\;} 
(a) GT  \;\;\;\;\;\;\;\;\;\;\;\;\;\;\;\;\;\;\;\;\;\;\;\;\;\;
(b) H-VQ-VAE  \;\;\;\;\;\;\;\;\;\;\;\;\;\;\;\;
(c) Attentive H-VQ-VAE  \;\;\;\;\;\;\;\;\;\;
(d) Attentive VQ-VAE  \;\;\;\; \\
\caption{(a) Random CelebA-HQ images $256\times256$ pixels, and their reconstructions computer with the proposed method: (b) Hierarchical without attention and two encoding levels; (c) Attentive--Hierarchical VQ-VAE  with two encoding levels; and (d) Attentive VQ-VAE with one encoding level.} 
\label{fig:reconstructions}
\end{figure*}

\mycomment{
\begin{figure}[!ht]
\centering
\includegraphics[width=\linewidth]{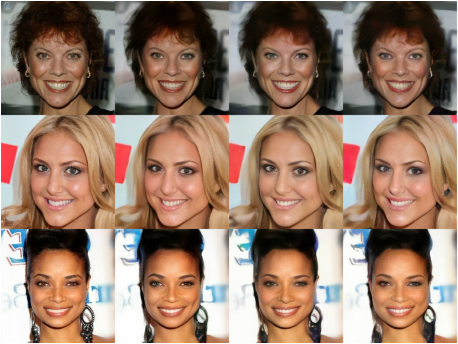}
\caption{A selection of reconstructed faces, the columns are in the same order than in Fig. \ref{fig:reconstructions}. The attentive model preserve better the face; compare the eyes.} 
\label{fig:reconstructions_details}
\end{figure}
}

\subsection{Implementation Details}
\label{ssec:implementation}

Assuming input images of $256 \times 256$ pixels with the color channels $(H,W,3)$, Base-Residual-Encoder transforms them to a tensor $(H,W,C)$

We based on ResNetV2 to implement our blocks "Id-ResBlock" and "Conv-ResBlock". Let us break down the structure of those blocks:

\begin{enumerate}
 \item Id-ResBlock (Identity Residual Block). Layers: BatchNormalization, LeakyReLU Activation (with a slope of $\alpha=0.1$), Convolution2D (with the number of filters equals the number of input channels), and the output of the last convolution is summed with the input to the block. Hence, this block's input and output have matching dimensions.
 
 \item Conv-ResBlock (Convolutional Residual Block). This block consists of the same layers as Id-ResBlock, with the difference that the Convolution2D has a stride equal to (2,2). With this stride, the width and height dimensions are reduced by half. For this reason, a Convolution2D with the same parameters is applied to the input data to be added to the output of the main execution path.

\end{enumerate}

In summary, both "Id-ResBlock" and "Conv-ResBlock" are designed to facilitate the flow of information through deep neural networks while addressing the vanishing gradient problem. The "Id-ResBlock" maintains input and output dimensions, while the "Conv-ResBlock" reduces dimensions through a stride in the convolutional layer and adjusts the input accordingly to allow for summation. These blocks are crucial for enabling the training of deep neural networks effectively. Table \ref{table:params} summarizes the reminder parameters; the encoders are of the kind residual with the distinction that the Base-Encoder ended with a $1 \times 1$ convolution with many filters equal to the latent dimension instead of an attention module of the ARENs. The Discriminator is a convolutional network with strides indicated in the next row of the table. 

\begin{table}[ht!]
 \footnotesize
  \centering
    \setlength\tabcolsep{4pt} 
  \begin{tabular}{lcccc}
 \hline
                &       & Convolutional &           \\
 Module         & R,r   & filters       &  Output    \\
 \hline\hline
 Base Res-Encoder & 3,2 & (128,128,128), +256  &  (64,64,266) \\
 AREN 1           & 2,2 & (128,128)            &   (32,32,256) \\
 AREN 2           & 3,2 & (128,128,128)        &   (16,16,256) \\
 AREN 3           & 3,2 & (128,128,128, 128)   &   (8,8,256) \\
 Discriminator    & --- & (128,128,128,64,64,1) &   (32,32,1) \\
 (strides)        & --- & (2,2,2,1,1,1)        &  \\
 \hline
\end{tabular}
\caption{\footnotesize Summary of model hyper-parameters.}
\label{table:params}
\end{table}

\vspace{-4mm}
\section{Experiments}
\label{sec:experiments}

\subsection{Comparative Results}
\label{saec:compare}

In this work, for demonstration purposes, we focus on face generation using the CelebA-HQ dataset\cite{karras2018progressive} with a resolution equal to $256 \times 256$ pixels. We used 80\% of the images for training and the remaining 20\% for testing.

Fig. \ref{fig:reconstructions} depicts generated faces from the test dataset; as we can see, these are very similar to the input faces. Panel (a) shows random examples of the test set (Ground True, GT). The following panels show the reconstructions computed with our proposal Attentive and Hierarchical VQ--VAE: 

\begin{itemize}
    \item Fig. \ref{fig:reconstructions}(b) Two levels of encoding without attention. The number of active vectors per level was [54,75] for the low and high levels. The predictions have a MAE$/\sigma = 0.2194$, where $\sigma=0.31$ is the variance of the data.

    \item Fig. \ref{fig:reconstructions}(c) Attentive Hierarchical VQ-VAE, the active vectors were [81,1] and  MAE$/\sigma=0.2138$.
    
    \item Fig. \ref{fig:reconstructions}(d) Attentive VQ-VAE using only one level, the active vectors were [55] and MAE$/\sigma=0.2259$.
    
\end{itemize}
  
We trained the models for 400 epochs in Nvidia 3090 RTX.
Our Hierarchical VQ-VAE (two-level) preserves the textures slightly better than the version with attention. However, the model with attention achieves better symmetry in the generated faces; we can note it when comparing the colors of both eyes. We noted that the attention mechanism alone was sufficient to incorporate long-range relationships between regions of the images. Therefore, we simplified the model by leaving a single level and the attention module, Fig \ref{fig:reconstructions}(d). The attentive model preserves better face symmetry.

Table \ref{table:resources} shows the computational resources demanded for each model. The patch-discriminator architecture ($0.412$ Millions of parameters) was the same for all the models. Since all the models were trained for the same epochs number, it is reasonable to expect that the AH-VQVAE model has not reached the same grade of convergence as the simplest model. That could explain some observed asymmetries.

\begin{table}[ht!]
 \footnotesize
  \centering
    \setlength\tabcolsep{4pt} 
  \begin{tabular}{ccc}
 \hline
                & Num. parameters & Training time  \\
 Model          & (millions)  & (secs. per epoch )      \\
 \hline\hline
 \,\,\,\,H-VQVAE        & 12.446 & 923   \\
 AH-VQVAE       & 12.496 & 1195   \\
 \,\,\,\,A-VQVAE        & {\bf 7.549} & {\bf 815}  \\
 \hline
\end{tabular}
\caption{\footnotesize Computational resources.}
\label{table:resources}
\end{table}

\subsection{Additional applications}
\label{saec:results}

\begin{figure}[!ht]
\centering
\includegraphics[width=0.3\linewidth]{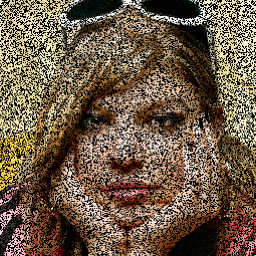}
\includegraphics[width=0.3\linewidth]{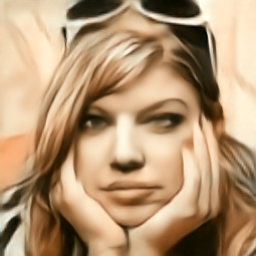}
\includegraphics[width=0.3\linewidth]{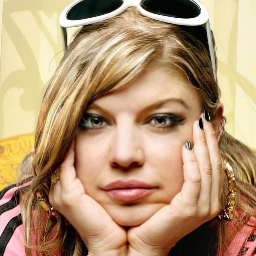}
\includegraphics[width=0.3\linewidth]{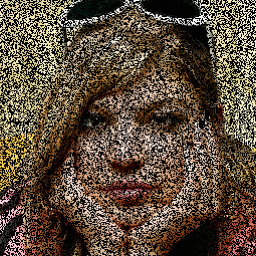}
\includegraphics[width=0.3\linewidth]{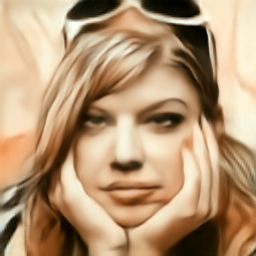}
\includegraphics[width=0.3\linewidth]{25076.png}
\includegraphics[width=0.3\linewidth]{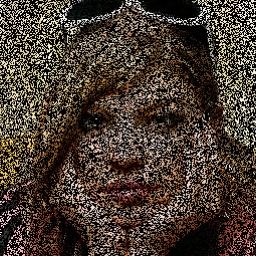}
\includegraphics[width=0.3\linewidth]{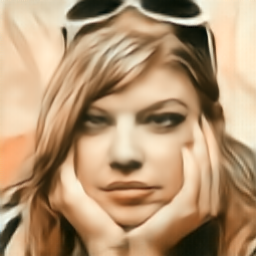}
\includegraphics[width=0.3\linewidth]{25076.png}
\caption{An example of test image in CelebA-HQ dataset with source image having blind mask effect (left), its Attentive VQ-VAE reconstructions (center) compared to the original image (right), with $40\%$ (up), $50\%$ (middle), $60\%$ (down) blind pixels.} 
\label{fig:blind}
\end{figure}

In the previous section, we show examples of how Attentive VQ-VAE can be used for face generation achieving impressive results. However, the Attentive VQ-VAE model has been designed to showcase its capabilities in diverse applications. One of these applications is blind restoration \cite{gu2022vqfr}, where some percent of the pixels in an image are lost, and the remaining pixels are used as input with a mask. The model is then trained to reconstruct the original image, with only partial information available. 
We train our model by randomly  masking as missed (with a uniform distribution) the 50\% of the pixels  and evaluate the trained model for masking the  $[30\%, 40\%, 50\% 60\%, 70\%]$ of the image pixels.  
This task can be challenging, but the Attentive VQ-VAE model can rebuild the image, with slight color changes, with a high precision. The results of this experiment can be seen in Fig. \ref{fig:blind}.

\begin{table}[!ht]
\centering
  \begin{tabular}{lcc}
    \hline
      \multicolumn{1}{c}{blind pixel} &
      \multicolumn{1}{c}{PSNR} &
      \multicolumn{1}{c}{SSIM} \\
    \hline\hline
    30\% & 21.333890 & 0.742172 \\
    40\% & 23.116606 & 0.763327 \\
    50\% (train) & 23.912767 & 0.765033 \\
    60\% & 23.309515 & 0.746209 \\
    70\% & 21.237381 & 0.693228 \\
    \hline
  \end{tabular}
  \caption{Peak Signal Noise Ratio (PSNR) and Structure Similarity Index (SSIM) in test CelebA-HQ dataset between original and reconstructed images with different percentages of blind pixels.}
    \label{table:blind}
\end{table}

\begin{figure}[!ht]
\centering
\includegraphics[width=0.3\linewidth]{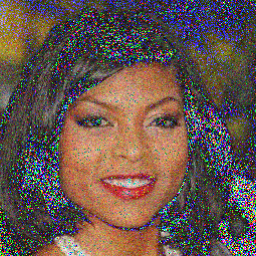}
\includegraphics[width=0.3\linewidth]{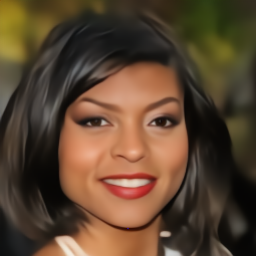}
\includegraphics[width=0.3\linewidth]{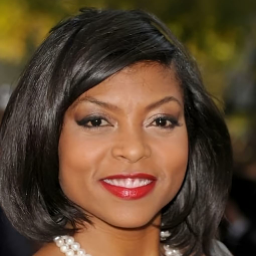}
\includegraphics[width=0.3\linewidth]{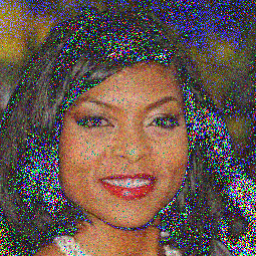}
\includegraphics[width=0.3\linewidth]{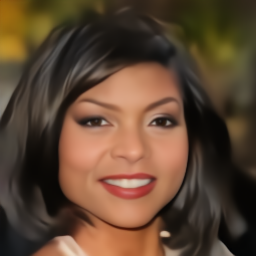}
\includegraphics[width=0.3\linewidth]{25960.png}
\includegraphics[width=0.3\linewidth]{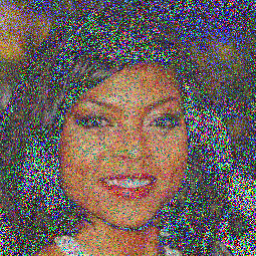}
\includegraphics[width=0.3\linewidth]{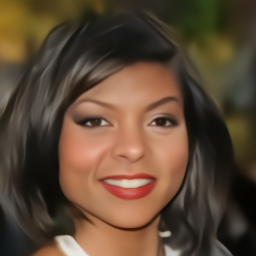}
\includegraphics[width=0.3\linewidth]{25960.png}
\caption{An example of test image in CelebA-HQ dataset with source image having gaussian noise (left), its Attentive VQ-VAE reconstructions (center) compared to the original image (right) with $\sigma=20\%$ (up), $30\%$ (middle), $40\%$ (down) Gaussian noise pixels.} 
\label{fig:denoise}
\end{figure}

Another classic reconstruction problem in image processing is denoising \cite{Peng_2021_CVPR}; in our case, the image pixels are corrupted by noise with a normal distribution, where the standard deviation of this noise is proportional to the image's dynamic range. 
We train our model adding Gaussian noise with standard deviation ($\sigma$) equal to the 30\% of the dynamic range of the image and evaluate the trained model for noisy data corrupted with Gaussian noise with $\sigma=\{20\%, 30\%, 40\% \}$. 
Fig. \ref{fig:denoise}, compares the original image and the image reconstructed by our Attentive VQ-VAE model. The reconstructed image appears to have different details in the necklace than the original, but this is the only noticeable difference between the two images. Overall, the reconstructed image is closely to the original, considering the presence of noise.

\begin{table}[!ht]
\centering
  \begin{tabular}{lcc}
    \hline
      \multicolumn{1}{c}{Noise $\sigma$} &
      \multicolumn{1}{c}{PSNR} &
      \multicolumn{1}{c}{SSIM} \\
    \hline\hline
    0.2 & 20.704477 & 0.7027168 \\
    0.3 (train) & 21.034735 & 0.7082805 \\
    0.4 & 18.611353 & 0.6841396 \\
    \hline
  \end{tabular}
  \caption{Peak Signal Noise Ratio (PSNR) and Structure Similarity Index (SSIM) in test CelebA-HQ dataset between original and reconstructed images with different percentages of noise pixels.}
\end{table}

\begin{figure}[!ht]
\centering
\includegraphics[width=0.3\linewidth]{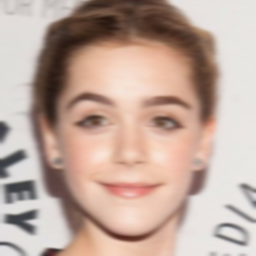}
\includegraphics[width=0.3\linewidth]{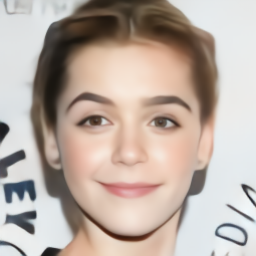}
\includegraphics[width=0.3\linewidth]{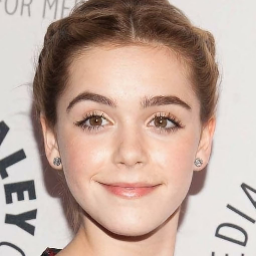}
\includegraphics[width=0.3\linewidth]{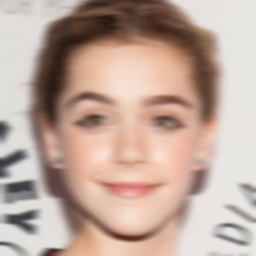}
\includegraphics[width=0.3\linewidth]{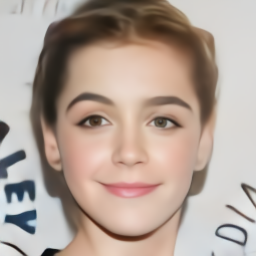}
\includegraphics[width=0.3\linewidth]{25252.png}
\includegraphics[width=0.3\linewidth]{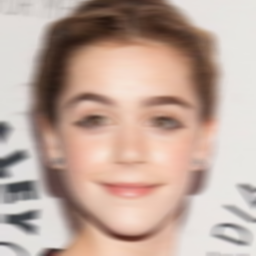}
\includegraphics[width=0.3\linewidth]{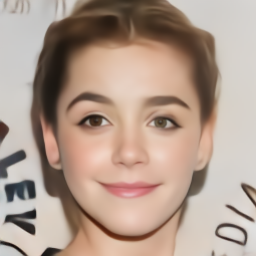}
\includegraphics[width=0.3\linewidth]{25252.png}
\caption{An example of test image in CelebA-HQ dataset with source image having blurring effect (left), its Attentive VQ-VAE reconstructions (center) compared to the original image (right) with $\sigma=3.0$ (up), $5.0$ (middle), $8.0$ (down) Gaussian noise pixels.} 
\label{fig:blurring}
\end{figure}

Finally, the deblurring problem \cite{sun2023degradation} refers to restoring a visually blurred images due to the motion effect, defocus, interpolation and is simulated by a  Gaussian filter (GF) with specific filter size and sigma values. 
In our example, we use a homogeneous GF with standar deviation ($\sigma$) equal 3.15 pixels and an additional directional GF (in the $y$ axis) with variance equal 5 pixels. For testing we evaluate the trained model variating the directional burring according with $\sigma=\{3.0, 5.0, 8.0\}$. Fig \ref{fig:blurring} shows examples of image reconstruction after deblurring the input. Once again, our proposed model did excellent work reconstructing the blurred image with only a slight change of color tone.

\begin{table}[!ht]
\centering
  \begin{tabular}{lcc}
    \hline
      \multicolumn{1}{c}{blurred $\sigma$} &
      \multicolumn{1}{c}{PSNR} &
      \multicolumn{1}{c}{SSIM} \\
    \hline\hline
    (1, 3) & 24.693070 & 0.82922704 \\
    (1, 5) (train) & 24.802713 & 0.78368074 \\
    (1, 8) & 26.407566 & 0.84432104 \\
    \hline
  \end{tabular}
  \caption{Peak Signal Noise Ratio (PSNR) and Structure Similarity Index (SSIM) in test CelebA-HQ dataset between original and reconstructed images with different percentages of blurred pixels.}
\end{table}

For each previous problem, we trained an Attentive VQ-VAE with one encoding level, latent space dimension 512, and 128 embeddings for 300 epochs on an RTX 3090. Tables 3-5 show the PSNR and SSIM metrics for blind restoration, denoising, and deblurring problems; respectively. In those tables, we can observe variations in the corruption parameters. This is because we observed that even when we trained our model with corrupted images with 50 percent of blind pixels, gaussian noise with $\mathcal{N}(0, \sigma=0.3)$ and filter size (3, 15) with sigma (1, 5) for the Gaussian filter, it was possible to add/remove some variations in this original values without retraining. Our model is robust enough to perform a good reconstruction, as shown in the metrics tables and example images in Fig. \ref{fig:blind}, \ref{fig:denoise}, \ref{fig:blurring}.

\vspace{-5mm}
\section{Conclusion}
\label{sec:conlcusion}
\vspace{-2mm}

We have introduced the Hierarchical Attentive VQ-VAE to enhance the capabilities of VQ-VAE models. The motivation behind this work arises from the need to address limitations in existing generative models, particularly those related to fine-grained details and global consistency in generated images. Our proposed Attentive VQ-VAE incorporates attention and hierarchical mechanisms. These additions aim to improve the encoding and generation capabilities of VQ-VAEs significantly.
Through experiments and developed applications, we demonstrated the effectiveness of the Attentive VQ-VAE in generating and restoring high-quality and realistic face images while simultaneously reducing the computational cost. Our attentive strategy exhibited similar performance in symmetry, color distribution, and facial feature preservation compared to a Hierarchical strategy. Furthermore, we used a strategy based on Generative Adversarial Networks (GANs) that contributes to more efficient and effective training. We confirm adding attention does not significantly increase the number of parameters, although it does increase the computational training time. On the other hand, we can obtain, through attention, results similar in quality to those obtained using a hierarchical strategy while keeping the complexity of the model and its training time under control.
In summary, our research showcases the potential of the Attentive VQ-VAE as a valuable tool for various applications, especially in image generation. Our model opens up exciting possibilities in computer vision, image processing, and generative art by addressing the challenges of fine-grained details and global consistency. Looking ahead, our work sets the stage for future research directions in generative modeling. Exploring the Attentive VQ-VAE's potential in other domains and extending its capabilities for even higher-resolution images represent promising avenues for further investigation. We believe this model can significantly contribute to the advancement of generative models and their applications in various creative and practical domains. Our models also implement a computational efficient hierarchical approach, future work will focus in developing applications that effectively combine the potential of both attention and hierarchical mechanisms.   

\vspace{2mm}
\noindent {\bf Acknowledges.} Work supported by CONAHCYT, Mexico (MR Grant CB-A1-43858, AH Scholarship). 



\bibliographystyle{IEEEbib}
\bibliography{rivera_attentive_vqvae}

\end{document}